\documentclass[runningheads]{llncs}

\usepackage{graphicx}
\usepackage{url}
\usepackage{times}
\usepackage{latexsym}
\usepackage[T1]{fontenc}
\usepackage[utf8]{inputenc}
\usepackage{microtype}
\usepackage{inconsolata}

\begin{document}

\title{Spot the bot: Coarse-Grained Partition of Semantic Paths for Bots and Humans}

\author{Vasilii A. Gromov\orcidID{0000-0001-5891-6597} \and
Alexandra S. Kogan\orcidID{0000-0002-6009-5203}}

\authorrunning{V. A. Gromov, A. S. Kogan}

\institute{National Research University Higher School of Economics,\\ 
Moscow, Russian Federation,\\
\email{stroller@rambler.ru}, \email{kogana00@gmail.com}}

\maketitle             

\begin{abstract}
Nowadays, technology is rapidly advancing: bots are writing comments, articles, and reviews. Due to this fact, it is crucial to know if the text was written by a human or by a bot. This paper focuses on comparing structures of the coarse-grained partitions of semantic paths for human-written and bot-generated texts. We compare the clusterizations of datasets of n-grams from literary texts and texts generated by several bots. The hypothesis is that the structures and clusterizations are different. Our research supports the hypothesis. As the semantic structure may be different for different languages, we investigate Russian, English, German, and Vietnamese languages.

\end{abstract}

\pdfoutput=1

\section{Introduction}

The Internet is a rich source of information. However, not all the information is trustworthy. There are special programs (bots) whose main aim is to write spam, malicious content, fake news, or reviews. Consequently, it is crucial to know whether a particular text was written by a human or a bot.

The majority of modern research deals with a particular bot, but, as far as we know, there is no in-depth research on the semantic paths of natural language texts. Coarse-grained partition of semantic paths may help find the structure of <<human>> language, and detect text generated by various bots.

A semantic path is defined as a sequence of embeddings of the words of the text. To gain a coarse-grained partition we split the semantic path into n-grams, concatenate the corresponding vectors and apply a clustering algorithm.

The main idea is that humans and bots write using different $n$-grams ($n > 1$ because bots can use the same set of words as humans). Humans can produce more complex and <<flowery>> phrases while bots use more standard phrases and may be repetitive. Therefore the clusterizations will differ.

\section{Related Works}

There are a lot of papers devoted to bot detection. Many of them use metadata or/and interaction of accounts (authors of the texts) \cite{lstm}, \cite{chak}, \cite{sent}, \cite{chu}.

However, the problem statement when bot detection is based purely on texts seems to be more practicable. Various neural networks, such as BiLSTM \cite{bilstm}, GPT \cite{gpt}, BERT \cite{bert}, DDN \cite{ddn}, etc. Graph Convolutional Networks are widely used for detecting bots in social media \cite{graph}, \cite{bert_graph}. Linguistic alignment can help detect bots in human-bot interactions \cite{alignment}.

In \cite{power_law} language was investigated as a whole and it was shown that  the number of words in the texts obeys power-law distribution, so it is possible to analyze <<human>> language not only on the level of particular texts.

In contrast to other research, our work is not focused on detecting a particular bot. We are trying to explore the structure of a natural language and use this structure to distinguish human-written and bot-generated texts.

\section{Methodology}

\subsection{Corpora}

We employ corpora of literary texts, because, from our point of view, literature is the clear reflection of the national language.

The methods for finding the structure of the language may differ for different languages or different families/branches, therefore we investigate several languages.

The majority of available datasets contain Wikipedia or news articles, blog posts, or only fragments of texts (the SVD embedding method requires full texts). Due to this fact, we  collect corpora ourselves from open sources, such as Project Gutenberg\footnote{\url{https://www.gutenberg.org/}}.

The preprocessing includes lemmatization and tokenization. Also, pronouns, proper nouns, numbers, and determiners (in German) are replaced with special tokens.  For preprocessing we use pretrained models\footnote{Russian \url{https://github.com/natasha/natasha}, English \url{https://spacy.io/models/en} (en\_core\_web\_lg), German \url{https://spacy.io/models/de} (de\_core\_news\_lg), Vietnamese \url{https://github.com/trungtv/pyvi}}.

The overview of corpora can be found in Table~\ref{tab:lang_corp}
\begin{table*}
\centering
\begin{tabular}{lllll}
\hline
\textbf{Language} & \textbf{Family}&\textbf{Branch} & \textbf{number of texts}&\textbf{avg. size of text, words}\\
\hline
Russian & Indo-European&Balto-Slavic & 12692 & 1000  \\
English & Indo-European&Germanic & 11008 & 21000  \\
German & Indo-European&Germanic & 2300 & 25000 \\
Vietnamese & Austroasiatic&Vietic & 1071 & 55000 \\

\hline
\end{tabular}
\caption{\label{tab:lang_corp}
Overview of the corpora
}
\end{table*}

\subsection{Bots}

We employ 2 bots for every language: LSTM (trained on our corpora) and GPT-2/3 (pretrained) \footnote{GPT-2 for Russian \url{https://huggingface.co/sberbank-ai/rugpt3large_based_on_gpt2}, English \url{https://huggingface.co/gpt2}, German \url{https://huggingface.co/dbmdz/german-gpt2} and GPT-3 for Vietnamese \url{https://huggingface.co/NlpHUST/gpt-neo-vi-small}}

To generate texts similar to texts in the corpus, every 100th word in the literary text is used as a prompt for the model and about 100 words are generated.

\subsection{Embeddings}

We employ SVD \cite{LSM} and Word2Vec\footnote{\url{https://radimrehurek.com/gensim/models/word2vec.html}} \cite{word2vec} embeddings. We train embeddings on our literary corpora.

The SVD embeddings rely on co-occurrences of the words in the texts, while Word2Vec ones (Skip-Gram, CBOW)  take into account the local context.

\subsection{Dataset creation}

To create a dataset we need:
\begin{itemize}
    \item $n$ - size of $n$-gram;
    \item $R$ - size of the embedding;
    \item A corpus of preprocessed texts;
    \item A dictionary (a word and corresponding vector) 
\end{itemize}

For every text in the corpus, all $n$-grams are investigated. For every word in the $n$-gram, the corresponding vector of size $R$ is taken from the dictionary. The vector for $n$-gram is the concatenation of vectors of its words. The resulting dataset contains vectors for $n$-grams, which have size $R \times n$.

In this research $n=2$, $R=8$

\subsection{Clustering}

The Wishart clustering technique\footnote{\url{https://github.com/Radi4/BotDetection/blob/master/Wishart.py}}\cite{wishart} was chosen during preliminary explorations (we compared several clustering techniques on synthetic datasets and dataset of literary texts). 

Calinski-Harabasz index\footnote{\url{https://scikit-learn.org/stable/modules/generated/sklearn.metrics.calinski_harabasz_score.html}} (CH) \cite{clustering} was chosen as a clustering metric in preliminary experiments. 

However, the clusterization chosen via CH may not always be useful because of the enormous number of noise points. To decrease it we adjuste the metric: $CH_{adj} = CH \times (ratio\_not\_noise)^T$, where $T$ is a hyperparameter, and $ratio\_not\_noise$ is number of point not marked as noise divided by total number of points.

\subsection{Comparing clusterizations}

To compare clusterizations of datasets corresponding to humans and bots, for every cluster metrics are computed. As a result, there are 2 arrays for every metric. The hypothesis that the arrays belong to the same distribution is tested.

We compute 8 metrics, their formulas and descriptions  can be found in Table \ref{tab:metrics_clusters}

\textbf{Notation}

$NC$ - number of clusters

$C_i$ - cluster $i$

$n_i$ - number of unique points in cluster $i$

$\overline{n}_i$ - total number of points (with duplicates) in cluster $i$

$c_i$ - centroid of cluster $i$

\begin{table*}
    \centering
    \caption{Metrics for clusters}
    \begin{tabular}{ l p{8.0cm}  l  }
    \textbf{№} &\textbf{description}& \textbf{formula} \\
\hline
1 & Number of unique vectors in the cluster & $\xi_i = \frac{n_i}{max_{j=1}^{NC} n_j}$  \\

  &Normalized by the size of the biggest cluster &  \\

2 &Number of unique vectors in the cluster.  &  $\xi_i = \frac{n_i}{\sum_{j=1}^{NC} n_j}$\\

& Normalized by the total number of unique vectors in the dataset. & \\

3 & Number of vectors with duplicates in the cluster   &  $\xi_i = \frac{\overline{n}_i}{max_{j=1}^{NC} \overline{n}_j}$ \\

&Normalized by the size (number of vectors with duplicates) of the biggest cluster & \\ 

4 & Number of vectors with duplicates in the cluster. &  $\xi_i = \frac{\overline{n}_i}{\sum_{j=1}^{NC} \overline{n}_j}$\\
&  Normalized by the total number of vectors with duplicates in the dataset. & \\

5 & Maximum distance to the centroid & $\xi_i = \max_{x \in C_i} d(x, c_i)$ \\ 

&& \\

6 & Average distance to the centroid & $\xi_i = \frac{\sum_{x \in C_i} d(x, c_i)}{n_i}$ \\ 
&& \\ 

7 & Maximum distance between points in the cluster & $\xi_i = \max_{x, y \in C_i, x\neq y} d(x, y)$ \\ 

 && \\ 

8 &Average distance between points in the cluster & $\xi_i = \frac{\sum_{x, y \in C_i, x\neq y} d(x, y)}{n_i(n_i-1)}$ \\ 
&& \\ \hline

\end{tabular}
    
    \label{tab:metrics_clusters}
\end{table*}

To test the hypotheses the Mann-Whitney-Wilcoxon test\footnote{\url{https://docs.scipy.org/doc/scipy/reference/generated/scipy.stats.mannwhitneyu.html}} is used \cite{criterion}. However, as the 8 hypotheses are tested, the multiple comparisons problem arises. To tackle it we employ the Holm-Bonferroni method\footnote{\url{https://www.statsmodels.org/dev/generated/statsmodels.stats.multitest.multipletests.html}} \cite{hol}.

\subsection{Taking the subset}

Clustering the big dataset ($>3$ million points) is time- and memory-consuming. Hyperparameter optimization will be impossible. To handle this problem we conduct preliminary experiments on literary data using nested datasets ($D_1 \subset D_2 \subset ... \subset D_l$) to find the significantly lesser dataset which will be representative. 

\section{Results}

For every language, we conduct separate experiments on all embeddings on our dataset (literary texts) to find the best number of unique points and the best $T$. See  Table~\ref{tab:params_dataset} for the final choices. .The numbers of unique points in datasets corresponding to bots are significantly lesser than in ones corresponding to humans, so the whole datasets are taken.

Let us stress that all the methods (Wishart technique, CH metric), and hyperparameters (R, n, T, size of the literary subsets) were chosen based on only literary data. The comparison with bot generated texts is the final step of the research.

\begin{table}
\centering
\begin{tabular}{lcc}
\hline
\textbf{Language} & \textbf{Number of unique points} & \textbf{T}\\\hline
Russian&$\approx 650$K&2 \\
English&$\approx 175$K&4 \\
German&$\approx 400$K&2 \\
Vietnamese&$\approx 700$K&2\\

\hline

\end{tabular}

\caption{The number of points in literary texts and $T$ for different languages}
\label{tab:params_dataset}
\end{table}

The results of the comparison can be found in Table~\ref{tab:final_res_all_l_all_e}. For particular language, embedding and bot there are 2 numbers in the cell: the number of metrics (out of 8) with significant differences  before and after correction for multiple testing. 

The Skip-Gram embedding seems to be insufficient for the task as it fails to distinguish the clusterization of humans and bots for the majority of languages. The SVD works fine in Russian,  and German, but is useless in the Vietnamese and English languages. The CBOW embedding shows the best results because it can spot the difference for all languages and all bots.

\begin{table*}
    \centering
    \caption{Results of comparison. For all languages, embeddings, and bots. }
    \begin{tabular}{lcccccccc}
\hline

Embedding \textbackslash  Language & \multicolumn{2}{c}{Russian} & \multicolumn{2}{c}{English}& \multicolumn{2}{c}{German} & \multicolumn{2}{c}{Vietnamese} \\

&LSTM&GPT2&LSTM&GPT2&LSTM&GPT2&LSTM&GPT3 \\ \hline
SVD & 5-3 & 5-4 & 2-0 & 3-0 & 4-4& 4-4&2-0&0-0 \\ 
\textbf{CBOW} & \textbf{3-1} &\textbf{ 6-4} & \textbf{8-8 }& \textbf{6-6} & \textbf{4-2} &\textbf{ 7-7}&\textbf{6-4}&\textbf{4-4} \\ 
Skip-Gram & 0-0 & 0-0 & 6-0 & 7-0 & 6-6 & 8-8&0-0&0-0 \\ \hline
\end{tabular}

    \label{tab:final_res_all_l_all_e}
\end{table*}

\begin{figure}[!h]
\centering
\includegraphics[scale=0.45]{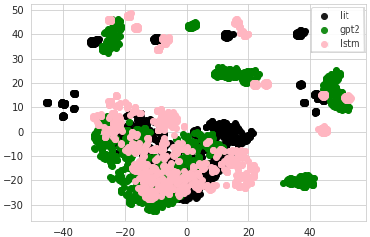}
\caption{t-SNE visualization of clusters. Language: Russian. Embedding: SVD}
\label{fig:russian_wishart_SVD_8_2_tsne}
\end{figure}

\begin{figure}[!h]
\centering
\includegraphics[scale=0.45]{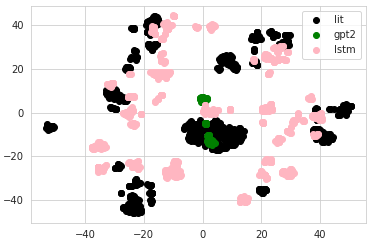}
\caption{t-SNE visualization of clusters. Language: Russian. Embedding: CBOW}
\label{fig:russian_wishart_cbow_8_2_tsne}
\end{figure}

The t-SNE\footnote{\url{https://scikit-learn.org/stable/modules/generated/sklearn.manifold.TSNE.html}} visualization \cite{tsne} for the Russian language the  SVD and CBOW embeddings can be seen in Figure~\ref{fig:russian_wishart_SVD_8_2_tsne} and Figure~\ref{fig:russian_wishart_cbow_8_2_tsne} correspondingly. The pink (light gray in grayscale), and green (dark gray) colors correspond to LSTM and GPT-2 bots and black is the color for humans (literature). It can be seen, that for the SVD method, there is a giant cluster for both bots and humans, and for the CBOW embedding there are a lot of little clusters. We also can notice that there are <<human>> clusters that are remote from bot clusters and vice versa, but t-SNE visualization  cannot provide enough evidence and additional experiments are required.

\section{Further research}

At this point, we have found differences between semantic paths for bots and humans while using a significant number of texts. We have not proposed a classifier for particular text yet. We plan to create a classifier with respect to results of this research.

We plan to investigate more advanced multilingual bots, such as mGPT \cite{mgpt} and more modern embeddings.

\section{Conclusion}

We can conclude that there are significant differences in coarse-grained partitions of semantic paths for bots and humans. We found them for all investigated languages and bots. The best results can be obtained using the CBOW embeddings.

\section*{Acknowledgments}
This research was supported in part through computational resources of HPC facilities at HSE University\cite{super}.

\bibliographystyle{splncs04} 

\bibliography{biblography}

\end{document}